%% file: main.tex
\documentclass{article}
\usepackage{spconf,amsmath,graphicx,hyperref}
\usepackage{microtype}
\usepackage{tikz}
\usepackage{pgfplots}
\pgfplotsset{compat=1.16}
\usepgfplotslibrary{groupplots}
\usepackage{amssymb}
\usepackage{float}
\usetikzlibrary{patterns}
\usepackage{booktabs}
\usepackage{multirow}
\definecolor{VisColor}{RGB}{55,138,221}
\definecolor{AudColor}{RGB}{31,158,117}
\definecolor{TxtColor}{RGB}{216,90,48}
\definecolor{MColor}{RGB}{20,125,146}
\definecolor{JointColor}{RGB}{212,84,60}
\newcommand{\pmsd}[1]{{\footnotesize$\pm$#1}}
\newcommand{\best}[1]{\textbf{#1}}
\newcommand{\second}[1]{\underline{#1}}

\newcommand{\maskrow}[5]{%
  \foreach \x in {0,...,6}{%
    \fill[#4] (#1+\x*0.24,#2) rectangle ++(0.21,0.25);%
  }%
  \foreach \x in {#5}{%
    \fill[gray!15] (#1+\x*0.24,#2) rectangle ++(0.21,0.25);%
    \fill[pattern=north east lines, pattern color=gray!65]
      (#1+\x*0.24,#2) rectangle ++(0.21,0.25);%
  }%
  \node[anchor=east,font=\scriptsize] at (#1-0.07,#2+0.125) {#3};%
}

\newcommand{\dashrule}[1]{%
  \noalign{\vskip\aboverulesep}%
  \multispan{#1}\leaders\hbox{\rule{5pt}{0.3pt}\hskip5pt}\hfill\\
  \noalign{\vskip\belowrulesep}}

\def\x{{\mathbf x}}

\title{TSMD: Temporal-Stream Modality Dropout for \\
Robust Video Highlight Detection}
\name{
  Bo-Yuan Cheng$^{1*}$, 
  Kuan-Yu Chen$^{1,2*}$, 
  Po-Han Huang$^{1}$, 
  Jeng-Lin Li$^{1}$,
  Jian-Jiun Ding$^{2}$%
  \thanks{* Equal contribution. Demo:
  \url{https://boyuan-ch.github.io/TSMD/}. Code:
  \url{https://github.com/boyuan-ch/TSMD}.}%
}
\address{
  $^{1}$AI Research Center, Inventec Corporation, Taiwan \\
  $^{2}$Graduate Institute of Communication Engineering, National Taiwan University, Taiwan
}

\begin{document}
\ninept
\maketitle
\begin{abstract}
Existing multimodal video highlight detectors typically assume that visual, audio, and textual streams are continuously available. In practice, however, inputs may suffer from localized frame missingness or complete-stream outage. We formulate this robustness challenge along two dimensions: temporal missingness, where frames are missing independently in each modality, and stream-level missingness, where one modality is unavailable throughout a video. Moreover, we find that the mean squared error (MSE) loss is misaligned with both the evaluation metrics and the peak-driven nature of highlights. Therefore, we propose \textbf{T}emporal-\textbf{S}tream \textbf{M}odality \textbf{D}ropout (\textbf{TSMD}), which combines structured missingness simulation with a joint objective comprising pointwise MSE, per-video Pearson correlation, and peak-oriented RankNet loss terms. TSMD has three variants: temporal, stream-level, and mixed dropout. On the MoSu and Mr.~HiSum datasets, TSMD-Temporal improves mAP@15 by $7.06$ and $3.41$ points over TripleSumm under $50\%$ independent temporal removal, whereas TSMD-Stream performs the best under complete-stream removal. TSMD-Mix retains most of these complementary benefits and ranks the best or the second-best across the evaluated temporal and stream-level conditions.
\end{abstract}

\input{section/1.Introduction}
\input{section/2.Method.tex}

\input{section/3.Experiments.tex}
\input{section/4.Conclusion.tex}

\clearpage
\bibliographystyle{IEEEbib}
\bibliography{refs}

\end{document}

%% file: section/1.Introduction.tex
\section{Introduction}
\label{sec:intro}

Video highlight detection identifies salient moments in untrimmed videos by predicting frame- or shot-level importance scores. Visual-only methods such as VASNet~\cite{fajtl2018vasnet} and CSTA~\cite{son2024csta} modeled visual sequences. Multimodal approaches incorporated audio or text in UMT~\cite{liu2022umt}, DAViHD~\cite{joo2026sounding}, and A2Summ~\cite{he2023a2summ} to resolve ambiguities that visual cues alone cannot address. TripleSumm~\cite{kim2026triplesumm} further models visual, audio, and text cues through adaptive fusion. This progression is reflected in the shift from small benchmarks such as SumMe~\cite{gygli2014summe} and TVSum~\cite{song2015tvsum} to large-scale datasets such as Mr.~HiSum~\cite{sul2023mrhisum} and the trimodal MoSu~\cite{kim2026triplesumm}.

\begin{figure}[t]
    \centering
    \resizebox{\columnwidth}{!}{\input{figs/fig1_deployment_teaser.tex}}
    \vspace{-18pt}
    \caption{Incomplete multimodal inputs in video highlight detection. Left:
    performance degrades under asynchronous \textbf{V}isual, \textbf{A}udio, and \textbf{T}ext corruption in deployment. Right: our retrained TripleSumm results on MoSu (Table~\ref{tab:main}). The missing-input panel
    reports $50\%$ independent temporal removal and mean complete-stream
    removal, with clean performance shown as a dashed reference.}
    \label{fig:teaser}
\end{figure}
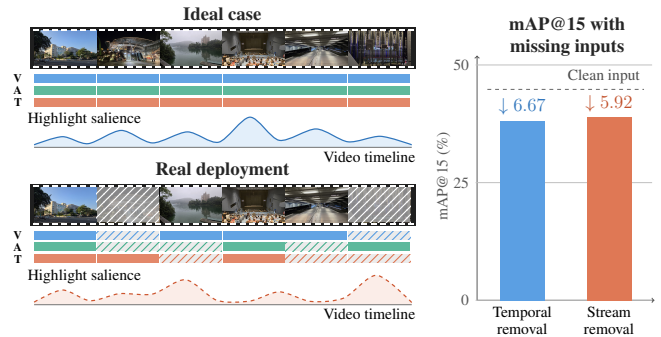

Existing multimodal detectors generally assume that all input streams are complete and uninterrupted. In highlight detection, this assumption is especially problematic. Unlike coarse video classification, where aggregate semantics dominate, salience prediction is fine-grained and peak-driven. Missing cues at key moments can suppress predicted scores and distort relative rankings throughout a video. Even TripleSumm~\cite{kim2026triplesumm}, which adaptively reweights modalities, loses $6.67$ and $5.92$ mAP@15 on MoSu under temporal and stream removal (Fig.~\ref{fig:teaser}). In real-world deployments, input quality is not guaranteed. At the frame level, network congestion can cause packet loss, poor video quality can degrade extracted visual features, and limitations of automatic speech recognition can produce incorrect transcripts. These localized, asynchronous issues can disrupt modality fusion and consequently degrade highlight detection. At the stream level, an unavailable audio track, a missing transcript, or a corrupted visual stream can effectively remove an entire modality for the full duration of a video. Robustness evaluation for video highlight detection should therefore consider both the temporal extent of feature corruption and which streams remain available.

Robustness to missing modalities has been studied mainly outside video highlight detection. ModDrop and EmbraceNet improve robust fusion for gesture recognition and general multimodal classification~\cite{neverova2016moddrop,choi2019embracenet}; MMIN and ActionMAE reconstruct missing modalities for emotion and action recognition~\cite{zhao2021mmin,woo2023actionmae}; and u-HuBERT uses masked multimodal speech pretraining~\cite{hsu2022uhubert}. Recent emotion and sentiment recognition studies further consider feature-level, temporal, and complete-modality missingness~\cite{zhong2025cider,yuan2024noise}. However, these studies neither target dense video highlight detection nor examine how temporal and complete-stream dropout policies interact during training. Different degradation processes also produce distinct missingness patterns. We therefore study both temporal and stream-level removal in dense highlight training and evaluation.

A separate challenge is the mismatch between pointwise training objectives and ranking-based highlight evaluation. Several baselines considered in this work, including VASNet, CSTA, and TripleSumm, apply pointwise mean squared error (MSE) to frame-level salience scores~\cite{fajtl2018vasnet,son2024csta,kim2026triplesumm}. Such supervision does not explicitly optimize relative ordering or top-segment retrieval. Prior highlight-oriented studies instead formulate ranking at the segment level: Yao et al.~\cite{yao2016pairwise} assign scores to video segments using pairwise deep ranking, while Mundnich et al.~\cite{mundnich2021audiovisual} predict one score per five-second clip and compare pointwise MSE with correlation- and margin-ranking losses. These studies operate on presegmented clips, whereas our setting requires dense frame-level salience prediction on a 1-Hz temporal grid. We therefore adopt a joint objective to balance performance across correlation-based and peak-oriented metrics.

To address these challenges, we introduce Temporal-Stream Modality Dropout (TSMD), a framework that combines structured missingness simulation with a joint objective. As a first step toward diverse real-world failures, we approximate missingness with a simple feature removal scheme. TSMD-Temporal simulates localized frame gaps, TSMD-Stream simulates complete-stream outages, and TSMD-Mix allocates corrupted training examples equally between these regimes to target both failure scales without adding parameters or inference computation. The joint objective complements pointwise MSE with per-video Pearson correlation for global salience-profile alignment and peak-oriented RankNet loss for highlight ordering. This design preserves clean-input performance while improving robustness. The main contributions of this work are:

\begin{itemize}\setlength{\itemsep}{1pt}\setlength{\parskip}{0pt}\setlength{\topsep}{2pt}
    \item A systematic formulation of multimodal missingness across temporal and stream dimensions, accompanied by a controlled benchmark evaluating independent temporal and complete-stream failures.
    \item A joint objective combining MSE, per-video Pearson correlation, and RankNet losses, which improves clean-input correlation and peak retrieval over the MSE-only baseline.
    \item Evaluations on MoSu and Mr. HiSum, showing that TSMD-Temporal gains up to $7.06$ mAP@15 over TripleSumm under $50\%$ temporal removal, while TSMD-Stream excels under stream removal.
    \item Evidence that temporal and stream robustness are complementary, with TSMD-Mix ranking the best or the second-best across the evaluated degradation conditions.
\end{itemize}

%% file: figs/fig1_deployment_teaser.tex
\begingroup
\begin{tikzpicture}[x=1cm,y=1cm]
\path[use as bounding box] (0.05,0.15) rectangle (17.65,9.40);

\node[font=\bfseries\Large,text=black!85] at (6.00,8.98) {Ideal case};
\node[font=\large,fill=white,inner sep=1pt] at (10.0,5.1) {Video timeline};

\fill[black!88] (0.78,7.48) rectangle (11.27,8.62);
\node[anchor=south west,inner sep=0pt] at (0.88,7.58)
  {\includegraphics[width=1.68cm,height=0.945cm]{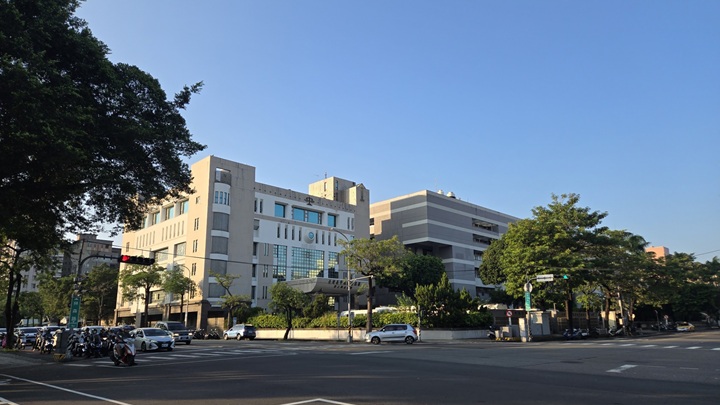}};
\node[anchor=south west,inner sep=0pt] at (2.59,7.58)
  {\includegraphics[width=1.68cm,height=0.945cm]{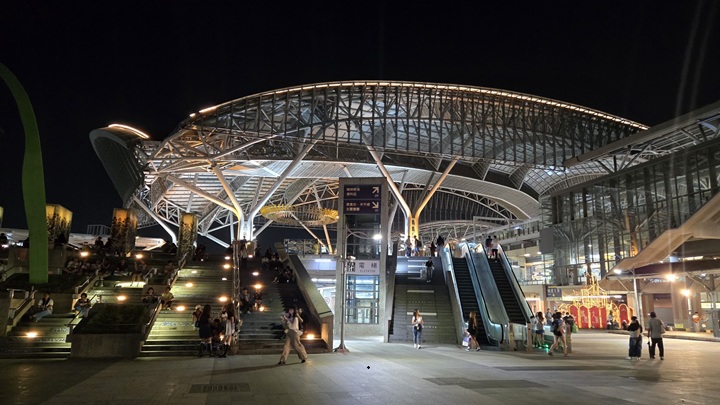}};
\node[anchor=south west,inner sep=0pt] at (4.30,7.58)
  {\includegraphics[width=1.68cm,height=0.945cm]{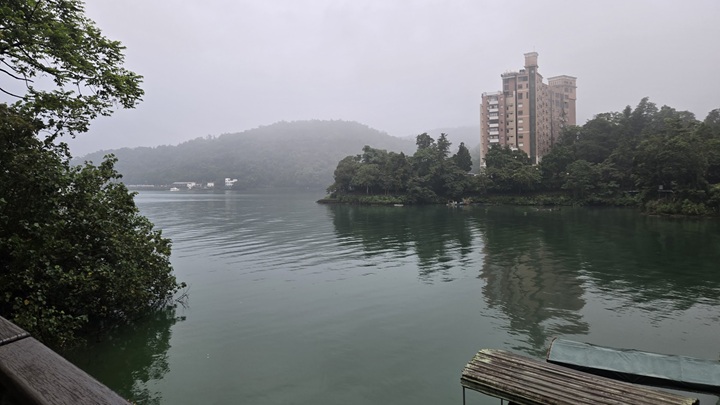}};
\node[anchor=south west,inner sep=0pt] at (6.01,7.58)
  {\includegraphics[width=1.68cm,height=0.945cm]{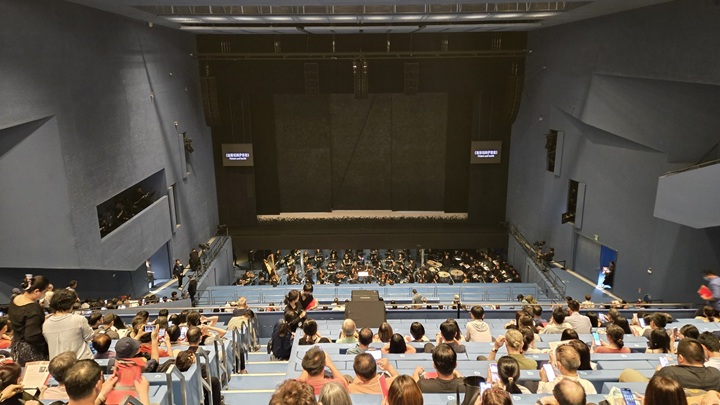}};
\node[anchor=south west,inner sep=0pt] at (7.72,7.58)
  {\includegraphics[width=1.68cm,height=0.945cm]{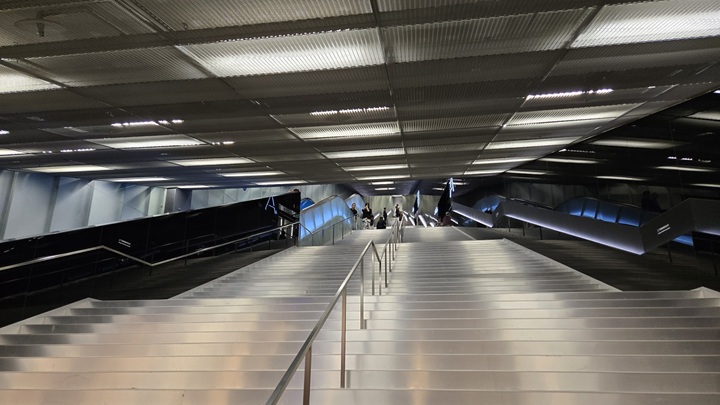}};
\node[anchor=south west,inner sep=0pt] at (9.43,7.58)
  {\includegraphics[width=1.68cm,height=0.945cm]{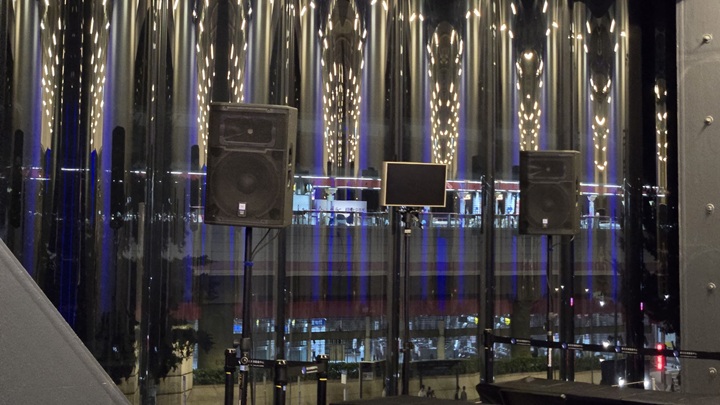}};
\foreach \x in {0,...,40}{
  \fill[white!92] ({0.87+0.255*\x},8.54) rectangle ++(0.14,0.045);
  \fill[white!92] ({0.87+0.255*\x},7.515) rectangle ++(0.14,0.045);
}

\node[anchor=east,font=\bfseries\small] at (0.66,7.23) {V};
\node[anchor=east,font=\bfseries\small] at (0.66,6.91) {A};
\node[anchor=east,font=\bfseries\small] at (0.66,6.59) {T};
\foreach \x in {0,...,5}{
  \fill[VisColor!78] ({0.88+1.71*\x},7.12) rectangle ++(1.68,0.22);
  \fill[AudColor!72] ({0.88+1.71*\x},6.80) rectangle ++(1.68,0.22);
  \fill[TxtColor!72] ({0.88+1.71*\x},6.48) rectangle ++(1.68,0.22);
}

\node[anchor=east,align=right,font=\large] at (3.9,6.12)
  {Highlight salience};
\fill[VisColor!14]
  plot[smooth] coordinates {(0.88,5.43) (1.65,5.65) (2.35,5.46)
  (3.25,5.82) (4.10,5.48) (5.05,5.78) (5.90,5.50)
  (6.75,6.18) (7.55,5.55) (8.55,5.86) (9.42,5.51)
  (10.30,5.66) (11.15,5.43)} -- (11.15,5.36) -- (0.88,5.36) -- cycle;
\draw[VisColor!85!black,line width=0.95pt]
  plot[smooth] coordinates {(0.88,5.43) (1.65,5.65) (2.35,5.46)
  (3.25,5.82) (4.10,5.48) (5.05,5.78) (5.90,5.50)
  (6.75,6.18) (7.55,5.55) (8.55,5.86) (9.42,5.51)
  (10.30,5.66) (11.15,5.43)};
\draw[black!55,line width=0.35pt] (0.88,5.36) -- (11.15,5.36);

\node[font=\bfseries\Large,text=black!85] at (6.00,4.72) {Real deployment};
\node[font=\large,fill=white,inner sep=1pt] at (10.0,0.84) {Video timeline};

\fill[black!88] (0.78,3.22) rectangle (11.27,4.36);
\node[anchor=south west,inner sep=0pt] at (0.88,3.32)
  {\includegraphics[width=1.68cm,height=0.945cm]{figs/1.jpg}};
\node[anchor=south west,inner sep=0pt] at (2.59,3.32)
  {\includegraphics[width=1.68cm,height=0.945cm]{figs/2.jpg}};
\node[anchor=south west,inner sep=0pt] at (4.30,3.32)
  {\includegraphics[width=1.68cm,height=0.945cm]{figs/3.jpg}};
\node[anchor=south west,inner sep=0pt] at (6.01,3.32)
  {\includegraphics[width=1.68cm,height=0.945cm]{figs/4.jpg}};
\node[anchor=south west,inner sep=0pt] at (7.72,3.32)
  {\includegraphics[width=1.68cm,height=0.945cm]{figs/5.jpg}};
\node[anchor=south west,inner sep=0pt] at (9.43,3.32)
  {\includegraphics[width=1.68cm,height=0.945cm]{figs/6.jpg}};
\fill[gray!65,opacity=0.78] (2.59,3.32) rectangle ++(1.68,0.945);
\fill[pattern=north east lines,pattern color=white!85]
  (2.59,3.32) rectangle ++(1.68,0.945);
\fill[gray!65,opacity=0.78] (9.43,3.32) rectangle ++(1.68,0.945);
\fill[pattern=north east lines,pattern color=white!85]
  (9.43,3.32) rectangle ++(1.68,0.945);
\foreach \x in {0,...,40}{
  \fill[white!92] ({0.87+0.255*\x},4.28) rectangle ++(0.14,0.045);
  \fill[white!92] ({0.87+0.255*\x},3.255) rectangle ++(0.14,0.045);
}

\node[anchor=east,font=\bfseries\small] at (0.66,2.97) {V};
\node[anchor=east,font=\bfseries\small] at (0.66,2.65) {A};
\node[anchor=east,font=\bfseries\small] at (0.66,2.33) {T};
\foreach \x in {0,...,5}{
  \fill[gray!16] ({0.88+1.71*\x},2.86) rectangle ++(1.68,0.22);
  \fill[gray!16] ({0.88+1.71*\x},2.54) rectangle ++(1.68,0.22);
  \fill[gray!16] ({0.88+1.71*\x},2.22) rectangle ++(1.68,0.22);
}
\foreach \x in {0,2,3,4}{
  \fill[VisColor!78] ({0.88+1.71*\x},2.86) rectangle ++(1.68,0.22);
}
\foreach \x in {0,3,5}{
  \fill[AudColor!72] ({0.88+1.71*\x},2.54) rectangle ++(1.68,0.22);
}
\foreach \x in {0,1,3}{
  \fill[TxtColor!72] ({0.88+1.71*\x},2.22) rectangle ++(1.68,0.22);
}
\foreach \x in {1,5}{
  \fill[pattern=north east lines,pattern color=VisColor!85]
    ({0.88+1.71*\x},2.86) rectangle ++(1.68,0.22);
}
\foreach \x in {1,2,4}{
  \fill[pattern=north east lines,pattern color=AudColor!85]
    ({0.88+1.71*\x},2.54) rectangle ++(1.68,0.22);
}
\foreach \x in {2,4,5}{
  \fill[pattern=north east lines,pattern color=TxtColor!85]
    ({0.88+1.71*\x},2.22) rectangle ++(1.68,0.22);
}

\node[anchor=east,align=right,font=\large] at (3.90,1.84)
  {Highlight salience};
\fill[TxtColor!11]
  plot[smooth] coordinates {(0.88,1.15) (1.65,1.48) (2.40,1.19)
  (3.20,1.38) (4.10,1.37) (5.00,1.76) (5.85,1.20)
  (6.75,1.18) (7.55,1.43) (8.40,1.17) (9.30,1.26)
  (10.20,1.88) (11.15,1.14)} -- (11.15,1.08) -- (0.88,1.08) -- cycle;
\draw[TxtColor!95!black,dashed,line width=0.95pt]
  plot[smooth] coordinates {(0.88,1.15) (1.65,1.48) (2.40,1.19)
  (3.20,1.38) (4.10,1.37) (5.00,1.76) (5.85,1.20)
  (6.75,1.18) (7.55,1.43) (8.40,1.17) (9.30,1.26)
  (10.20,1.88) (11.15,1.14)};
\draw[black!55,line width=0.35pt] (0.88,1.08) -- (11.15,1.08);

\node[font=\bfseries\Large,text=black!85,align=center] at (15.38,8.38)
  {mAP@15 with\\missing inputs};
\begin{scope}[xshift=15]
\draw[->,black!65,line width=0.45pt] (12.40,1.20) -- (12.40,7.85);
\draw[->,black!65,line width=0.45pt] (12.40,1.20) -- (17.22,1.20);
\foreach \value/\label in {0/0,25/25,50/50}{
  \draw[black!25,line width=0.25pt]
    (12.36,{1.20+0.128*\value}) -- (17.16,{1.20+0.128*\value});
  \node[anchor=east,font=\large,text=black!65]
    at (12.29,{1.20+0.128*\value}) {\label};
}
\node[anchor=south,rotate=90,font=\large,text=black!65]
  at (11.78,4.58) {mAP@15 (\%)};

\draw[black!60,dashed,line width=0.65pt]
  (12.68,{1.20+0.128*44.81}) -- (16.94,{1.20+0.128*44.81});
\node[anchor=south east,font=\large,text=black!70]
  at (16.94,{1.25+0.128*44.81}) {Clean input};
\fill[VisColor!82]
  (13.03,1.20) rectangle (14.23,{1.20+0.128*38.14});
\fill[TxtColor!82]
  (15.39,1.20) rectangle (16.59,{1.20+0.128*38.89});

\node[font=\bfseries\Large,anchor=south,text=VisColor!85!black]
  at (13.63,{1.35+0.128*38.14-0.1}) {$\downarrow 6.67$};
\node[font=\bfseries\Large,anchor=south,text=TxtColor!90!black]
  at (15.99,{1.35+0.128*38.89-0.1}) {$\downarrow 5.92$};
\node[font=\large,align=center] at (13.63,0.66) {Temporal\\removal};
\node[font=\large,align=center] at (15.99,0.66) {Stream\\removal};
\end{scope}

\end{tikzpicture}
\endgroup

%% file: section/2.Method.tex
\section{Method}
\label{sec:method}

\begin{figure}[t]
\centering
\begin{tikzpicture}
\begin{scope}[xshift=-10]
\draw[rounded corners=1.5pt, line width=0.5pt] (0.10,3.55) rectangle (0.90,4.25);
\fill[black] (0.40,3.75) -- (0.40,4.05) -- (0.67,3.90) -- cycle;
\node[font=\normalsize] at (0.50,3.37) {Video};

\draw[rounded corners=2pt, fill=VisColor!18, draw=VisColor!85!black]
    (1.48,4.25) rectangle (2.38,4.55);
\node[font=\normalsize] at (1.93,4.40) {Visual};
\draw[rounded corners=2pt, fill=AudColor!18, draw=AudColor!85!black]
    (1.48,3.75) rectangle (2.38,4.05);
\node[font=\normalsize] at (1.93,3.90) {Audio};
\draw[rounded corners=2pt, fill=TxtColor!18, draw=TxtColor!85!black]
    (1.48,3.25) rectangle (2.38,3.55);
\node[font=\normalsize] at (1.93,3.40) {Text};
\draw[->, line width=0.55pt] (0.90,3.90) -- (1.15,3.90)
    -- (1.15,4.40) -- (1.48,4.40);
\draw[->, line width=0.55pt] (0.90,3.90) -- (1.48,3.90);
\draw[->, line width=0.55pt] (0.90,3.90) -- (1.15,3.90)
    -- (1.15,3.40) -- (1.48,3.40);

\draw[rounded corners=2pt, fill=gray!18, draw=gray!65,
            line width=0.55pt] (2.90,3.52) rectangle (4.05,4.28);
\node[font=\normalsize,align=center] at (3.475,3.90) {Dropout\\mask};
\draw[->, line width=0.45pt] (2.38,4.40) -- (2.65,4.40)
    -- (2.65,4.08) -- (2.90,4.08);
\draw[->, line width=0.45pt] (2.38,3.90) -- (2.90,3.90);
\draw[->, line width=0.45pt] (2.38,3.40) -- (2.65,3.40)
    -- (2.65,3.72) -- (2.90,3.72);

\draw[rounded corners=2pt, fill=JointColor!14, draw=JointColor!80!black,
            line width=0.55pt] (4.39,3.52) rectangle (5.74,4.28);
\node[font=\normalsize,align=center] at (5.065,3.90) {Highlight\\detector};
\draw[->, line width=0.75pt] (4.05,3.90) -- (4.39,3.90);

\draw[->, gray!75, line width=0.4pt] (6.15,3.30) -- (6.15,4.48);
\draw[->, gray!75, line width=0.4pt] (6.15,3.30) -- (8.20,3.30);
\draw[JointColor!90!black, line width=0.9pt, smooth]
    plot coordinates {(6.15,3.47) (6.32,3.60) (6.49,3.52) (6.66,3.80)
    (6.83,3.61) (7.00,4.28) (7.17,3.68) (7.34,3.55)
    (7.51,3.84) (7.68,3.62) (7.85,4.34) (8.07,3.66)};
\node[anchor=west,font=\normalsize] at (6.17,4.51) {Predicted salience};
\node[anchor=east,font=\normalsize,text=gray!75] at (8.20,3.13) {time};
\draw[->, line width=0.75pt] (5.74,3.90) -- (6.08,3.90);


\draw[gray!65, line width=0.65pt] (3.475,3.52) -- (3.475,3.02);
\end{scope}

\draw[gray!65, line width=0.45pt] (1.85,3.02) -- (6.35,3.02);
\draw[->, gray!65, line width=0.55pt] (1.85,3.02) -- (1.85,2.75);
\draw[->, gray!65, line width=0.55pt] (6.35,3.02) -- (6.35,2.75);

\node[font=\normalsize] at (1.85,2.52) {(a) Independent temporal dropout};
\begin{scope}[xscale=1.25,yshift=3]
\maskrow{0.655}{1.78}{V}{VisColor}{0,3,6}
\maskrow{0.655}{1.47}{A}{AudColor}{1,4}
\maskrow{0.655}{1.16}{T}{TxtColor}{2,5}
\draw[->, gray!70, line width=0.4pt] (0.655,0.98) -- (2.305,0.98);
\end{scope}

\node[font=\normalsize] at (6.35,2.52) {(b) Complete-stream dropout};
\begin{scope}[xscale=1.25,yshift=3]
\maskrow{4.255}{1.78}{V}{VisColor}{0,...,6}
\maskrow{4.255}{1.47}{A}{AudColor}{}
\maskrow{4.255}{1.16}{T}{TxtColor}{}
\draw[->, gray!70, line width=0.4pt] (4.255,0.98) -- (5.905,0.98);
\end{scope}

\begin{scope}[yshift=3]
\node[anchor=west,font=\normalsize,text=gray!75] at (0.819,0.76) {time};
\node[anchor=west,font=\normalsize,text=gray!75] at (5.319,0.76) {time};
\fill[gray!15] (6.45,0.65) rectangle ++(0.21,0.17);
\fill[pattern=north east lines, pattern color=gray!65]
    (6.45,0.65) rectangle ++(0.21,0.17);
\node[anchor=west,font=\normalsize] at (6.74,0.76) {masked};
\end{scope}

\end{tikzpicture}
\vspace{-15pt}
\caption{Multimodal highlight detection and TSMD masking patterns:
(a) independent temporal and (b) complete-stream
dropout, illustrated for the visual stream.}
\label{fig:masking}
\vspace{-5pt}
\end{figure}
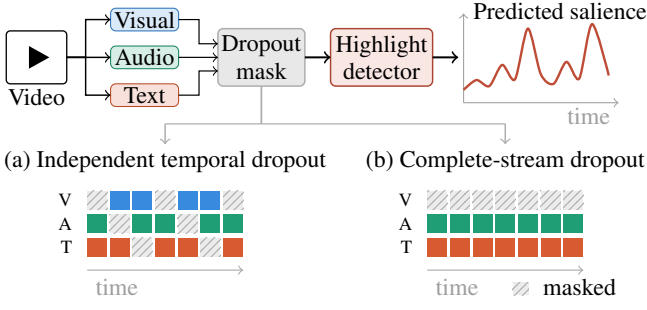


We formulate video highlight detection as a sequence-to-sequence regression problem. Let $N$ denote the number of temporally aligned timesteps in a video, with $n \in \{1,\ldots,N\}$ indexing a timestep. 
Let $X^m \in \mathbb{R}^{N\times d_m}$ denote the feature sequence of modality $m \in \{V, A, T\}$ (visual, audio, text), where $d_m$ is its feature dimension.
A highlight detector $f_\theta$ with trainable parameters $\theta$ maps the aligned feature sequences to a frame-level salience-score vector:
\begin{equation}
\hat{\mathbf{y}} = f_\theta(X^V, X^A, X^T)
= (\hat y_1,\ldots,\hat y_N) \in [0,1]^N .
\end{equation}
Here, $\hat y_n$ is the predicted salience score at timestep $n$. The corresponding ground-truth vector is $\mathbf y=(y_1,\ldots,y_N)\in[0,1]^N$, where $y_n$ is the ground-truth salience score at the same timestep.

\begin{table*}[t]
\centering
\caption{Results on MoSu and Mr.~HiSum datasets under \textbf{(a)} clean inputs, \textbf{(b)} independent temporal removal ($r{=}0.5$), and \textbf{(c)} complete-stream removal (averaged across $-$V/$-$A/$-$T). Published baselines from~\cite{kim2026triplesumm} appear above the dashed line in (a). All other rows are our three-seed mean $\pm$ SD. TripleSumm$^\dagger$ denotes our retrained MSE baseline, and all TSMD variants use the joint objective. The \textbf{best} and \second{second best} are marked per column; $\tau$ and $\rho$ are ranked at full precision. Mod.: input modalities.}
\label{tab:main}
\setlength{\tabcolsep}{3.2pt}
\begin{tabular}{@{}l c cccc cccc@{}}
\toprule
 & & \multicolumn{4}{c}{\textbf{MoSu}} & \multicolumn{4}{c}{\textbf{Mr.~HiSum}} \\
\cmidrule(lr){3-6}\cmidrule(lr){7-10}
Method / training policy & Mod. & $\tau\uparrow$ & $\rho\uparrow$ & mAP@50$\uparrow$ & mAP@15$\uparrow$ & $\tau\uparrow$ & $\rho\uparrow$ & mAP@50$\uparrow$ & mAP@15$\uparrow$ \\
\midrule
\multicolumn{10}{@{}l}{\textit{(a) Clean inputs}} \\
VASNet~\cite{fajtl2018vasnet}          & V   & 0.151 & 0.219 & 64.49 & 31.05 & 0.069 & 0.102 & 58.69 & 25.28 \\
A2Summ~\cite{he2023a2summ}             & VT  & 0.181 & 0.257 & 66.48 & 35.70 & 0.121 & 0.172 & 63.20 & 32.34 \\
UMT~\cite{liu2022umt}                  & VA  & 0.239 & 0.334 & 68.83 & 36.73 & 0.178 & 0.253 & 66.81 & 35.65 \\
CSTA~\cite{son2024csta}                & V   & 0.291 & 0.398 & 71.77 & 40.65 & 0.128 & 0.185 & 63.38 & 30.42 \\
TripleSumm~\cite{kim2026triplesumm}    & VAT & 0.351 & 0.472 & 74.72 & 44.42 & 0.258 & 0.352 & 70.72 & 40.88 \\

\dashrule{10}

TripleSumm$^\dagger$ & VAT & 0.353\pmsd{0.002} & 0.473\pmsd{0.002} & 74.73\pmsd{0.12} & 44.81\pmsd{0.11} & 0.255\pmsd{0.005} & 0.347\pmsd{0.007} & 70.47\pmsd{0.36} & 41.12\pmsd{0.21} \\
TSMD-Stream & VAT & \best{0.364}\pmsd{0.002} & \best{0.486}\pmsd{0.002} & \best{75.32}\pmsd{0.12} & \best{45.91}\pmsd{0.20} & \best{0.265}\pmsd{0.003} & \best{0.360}\pmsd{0.003} & \best{70.90}\pmsd{0.14} & \best{42.21}\pmsd{0.29} \\
TSMD-Temporal & VAT & 0.361\pmsd{0.002} & 0.483\pmsd{0.002} & 75.14\pmsd{0.17} & 45.65\pmsd{0.24} & 0.259\pmsd{0.006} & 0.351\pmsd{0.009} & 70.66\pmsd{0.44} & 42.06\pmsd{0.26} \\
TSMD-Mix & VAT & \second{0.364}\pmsd{0.003} & \second{0.485}\pmsd{0.003} & \second{75.31}\pmsd{0.17} & \second{45.89}\pmsd{0.36} & \second{0.262}\pmsd{0.005} & \second{0.355}\pmsd{0.007} & \second{70.74}\pmsd{0.33} & \second{42.19}\pmsd{0.05} \\

\midrule

\multicolumn{10}{@{}l}{\textit{(b) Missing frame} --- independent temporal removal, $r{=}0.5$ } \\
TripleSumm$^\dagger$ & VAT & 0.251\pmsd{0.014} & 0.347\pmsd{0.019} & 69.63\pmsd{0.90} & 38.14\pmsd{1.53} & 0.179\pmsd{0.019} & 0.252\pmsd{0.024} & 66.62\pmsd{0.52} & 38.19\pmsd{1.67} \\
TSMD-Stream & VAT & 0.314\pmsd{0.004} & 0.420\pmsd{0.006} & 72.68\pmsd{0.19} & 42.46\pmsd{0.24} & 0.221\pmsd{0.012} & 0.304\pmsd{0.015} & 68.58\pmsd{0.75} & 39.07\pmsd{0.15} \\
TSMD-Temporal & VAT & \second{0.354}\pmsd{0.003} & \second{0.474}\pmsd{0.003} & \second{74.85}\pmsd{0.25} & \second{45.20}\pmsd{0.23} & \second{0.251}\pmsd{0.005} & \second{0.341}\pmsd{0.007} & \second{70.28}\pmsd{0.31} & \best{41.60}\pmsd{0.14} \\
TSMD-Mix & VAT & \best{0.354}\pmsd{0.003} & \best{0.474}\pmsd{0.004} & \best{74.86}\pmsd{0.23} & \best{45.25}\pmsd{0.26} & \best{0.253}\pmsd{0.005} & \best{0.345}\pmsd{0.006} & \best{70.35}\pmsd{0.32} & \second{41.47}\pmsd{0.09} \\

\midrule

\multicolumn{10}{@{}l}{\textit{(c) Missing modality} --- complete-stream removal, mean over $-$V/$-$A/$-$T } \\
TripleSumm$^\dagger$ & VAT & 0.269\pmsd{0.009} & 0.366\pmsd{0.013} & 70.10\pmsd{0.47} & 38.89\pmsd{0.30} & 0.172\pmsd{0.009} & 0.236\pmsd{0.012} & 66.15\pmsd{0.48} & 37.81\pmsd{0.11} \\
TSMD-Stream & VAT & \best{0.332}\pmsd{0.002} & \best{0.446}\pmsd{0.002} & \best{73.49}\pmsd{0.09} & \best{43.30}\pmsd{0.11} & \best{0.239}\pmsd{0.002} & \best{0.326}\pmsd{0.002} & \best{69.50}\pmsd{0.06} & \best{40.40}\pmsd{0.37} \\
TSMD-Temporal & VAT & 0.312\pmsd{0.003} & 0.420\pmsd{0.003} & 72.40\pmsd{0.18} & 42.34\pmsd{0.10} & 0.216\pmsd{0.004} & 0.294\pmsd{0.005} & 68.32\pmsd{0.20} & 39.32\pmsd{0.55} \\
TSMD-Mix & VAT & \second{0.328}\pmsd{0.002} & \second{0.440}\pmsd{0.003} & \second{73.31}\pmsd{0.14} & \second{43.06}\pmsd{0.29} & \second{0.233}\pmsd{0.004} & \second{0.317}\pmsd{0.005} & \second{69.22}\pmsd{0.32} & \second{40.22}\pmsd{0.29} \\
\bottomrule
\end{tabular}
\vspace{-10pt}
\end{table*}

\subsection{Metric-Aligned Training Objective}
\label{ssec:objective}
Most-replayed annotations~\cite{kim2026triplesumm,sul2023mrhisum} are normalized to $[0,1]$ within each video and thus primarily encode relative salience. The conventional MSE loss is:
\begin{equation}
L_M = \frac{1}{N}\sum_{n=1}^{N}(\hat y_n - y_n)^2.
\end{equation}
It enforces pointwise magnitude matching, which may underemphasize the global profile. It also does not explicitly supervise relative frame ordering.
We therefore incorporate a per-video Pearson correlation loss to align salience profiles independently of shift and positive scaling:
\begin{equation}
L_P = 1 - r_{\mathrm P}(\hat{\mathbf y},\mathbf y),
\end{equation}
where $r_{\mathrm P}(\hat{\mathbf y},\mathbf y)$ denotes the Pearson correlation coefficient between the predicted and ground-truth salience vectors. Together, $L_M$ and $L_P$ align pointwise scores and the global salience profile. Because highlight detection centers on identifying the most important moments, we further employ RankNet~\cite{burges2005ranknet} to separate top highlights from other frames through ordinal supervision. Let $\mathbf z=(z_1,\ldots,z_N)\in\mathbb R^N$ denote the network's pre-sigmoid logits, such that $\hat y_n=\sigma(z_n)$. Specifically, we sample 512 candidate frame-index pairs $(i,j)$ per video, 
where frame $i$ is drawn from the top-$15\%$ ground-truth frames and frame $j$ is drawn from all frames. We retain pairs satisfying $|y_i-y_j|\ge0.1$ and denote the resulting pair set by $\mathcal P$. Given the target relation $s_{ij} = \operatorname{sign}(y_i - y_j)$, the ranking loss is formulated as:
\begin{equation}
L_R = \frac{1}{|\mathcal P|}\sum_{(i,j)\in\mathcal P} \log\!\left(1 + e^{-s_{ij}(z_i - z_j)}\right).
\end{equation}
$L_R$ emphasizes the correct ordering around key highlight frames, complementing the pointwise and profile-level supervision. The overall objective combines these three terms with weights set by preliminary experiments:
\begin{equation}
L_{\mathrm{MPR}} = L_M + 0.35L_P + 0.10L_R.
\end{equation}
TSMD variants share this objective, differing only in dropout policy.

\subsection{Temporal and Stream-Level Modality Removal}
\label{ssec:corruption}
Following ModDrop~\cite{neverova2016moddrop}, we implement the feature removal scheme by zero-masking. We formulate this as a unified masking operator, which corrupts inputs during TSMD training (Sec. \ref{ssec:loss}) and constructs the robustness benchmark at evaluation (Sec. \ref{ssec:evaluation}). Let $x_{n,m}\in\mathbb{R}^{d_m}$ denote the encoder feature vector at timestep $n$ and modality $m \in \{V,A,T\}$. An availability indicator $a_{n,m} \in \{0,1\}$ produces the corrupted feature
\begin{equation}
\widetilde{x}_{n,m} = a_{n,m}x_{n,m}.
\end{equation}
The selected feature vectors are set to zero, while sequence length, timestamps, targets, and the padding mask remain unchanged. This preserves positional encodings and frame--target alignment on the fixed temporal grid.

We simulate two forms of missingness (Fig.~\ref{fig:masking}). For \emph{temporal removal}, availability varies across timesteps: given a masking ratio $r$, exactly $k = \operatorname{round}(rN)$ timesteps are zeroed per stream. Masking is applied independently across modalities. For \emph{complete-stream removal}, one modality is zeroed across all timesteps.

\subsection{Temporal-Stream Modality Dropout (TSMD)}
\vspace{-5pt}
\label{ssec:loss}
TSMD introduces three training augmentations that simulate complementary missingness patterns while maintaining an overall $25\%$ clean-data exposure rate. 

\textbf{TSMD-Temporal} applies independent temporal dropout: for each video and epoch, a masking ratio $r$ shared by all streams is sampled uniformly from $\{0,\allowbreak 0.1,\allowbreak 0.3,\allowbreak 0.5\}$, and frames are masked independently across streams. $r=0$ yields the $25\%$ clean training instances, whereas $r>0$ introduces scattered temporal gaps. 

\textbf{TSMD-Stream} applies complete-stream dropout: each training instance is uniformly assigned a state from $\{\mathrm{clean}, -V, -A, -T\}$, where $-m$ denotes zeroing modality $m$ at all timesteps. This yields $25\%$ clean instances and $75\%$ single-stream outages. Multiple modalities are never dropped simultaneously, so at least two modalities remain for cross-modal fusion.

\textbf{TSMD-Mix} combines both failure modes through three mutually exclusive states: $25\%$ clean inputs, $37.5\%$ temporal dropout with $r$ sampled uniformly from$\{0.1, 0.3, 0.5\}$, and $37.5\%$ stream dropout with the missing stream sampled from $\{V,A,T\}$. None of the variants requires an auxiliary head, architectural modification, or additional inference computation.

%% file: section/3.Experiments.tex
\vspace{-5pt}
\section{Experiments}

\label{sec:exp}


\subsection{Datasets and Base Model}
\vspace{-5pt}

We evaluate on two large-scale benchmarks: \textbf{MoSu}~\cite{kim2026triplesumm} and \textbf{Mr.~HiSum}~\cite{sul2023mrhisum}. MoSu contains $52{,}678$ videos with aligned visual, audio, and transcript-based text streams, and an official train/validation/test split of $42{,}152$/\allowbreak$5{,}263$/\allowbreak$5{,}263$. Mr.~HiSum originally provides only visual features for $31{,}892$ YouTube videos with most-replayed statistics. We use the trimodal version released by~\cite{kim2026triplesumm}, which adds audio features from the raw videos and text features from generated frame captions, and retains the $30{,}452$ videos that remain accessible, split into $26{,}639$/\allowbreak$1{,}904$/\allowbreak$1{,}909$.

We adopt TripleSumm~\cite{kim2026triplesumm} as the backbone and leave its architecture unchanged. For both datasets, we use the pre-extracted features released by~\cite{kim2026triplesumm}. Audio and text features are obtained from pretrained AST~\cite{gong2021ast} and RoBERTa~\cite{liu2019roberta} encoders. Visual features are obtained from CLIP~\cite{radford2021clip} for MoSu and are the PCA-reduced InceptionV3~\cite{szegedy2016rethinking} features of~\cite{sul2023mrhisum} for Mr.~HiSum, yielding $(d_V,d_A,d_T)=(768,768,768)$ and $(1024,768,768)$, respectively. TripleSumm projects each modality into a shared embedding space, models intra-modal temporal context using Multi-scale Temporal blocks with progressively expanding self-attention windows, and performs timestep-wise multimodal fusion through interleaved Cross-modal Fusion blocks.

Models are optimized using AdamW (learning rate $10^{-4}$, weight decay $10^{-5}$) with cosine decay, batch size $64$, and early stopping (patience $10$, up to $100$ epochs) across three random seeds ($42$, $123$, $2026$). Checkpoints are selected by validation $\tau + \rho$ on uncorrupted videos.

\vspace{-5pt}
\subsection{Evaluation Metrics and Missingness Benchmark}
\label{ssec:evaluation}
\vspace{-5pt}
We evaluate sequence alignment via Kendall's $\tau$ and Spearman's $\rho$, and segment retrieval via mAP@50 and mAP@15~\cite{sul2023mrhisum,otani2019rethinking}. 
Kendall's $\tau$ measures pairwise order consistency among frames, whereas Spearman's $\rho$ measures the agreement between the predicted and ground-truth frame ranks.
For mAP, each video is partitioned into non-overlapping five-second segments, each scored by averaging its frame predictions. Segments ranked in the top 50\% or 15\% by ground-truth score are treated as positives for mAP@50 and mAP@15, which evaluate broader summary retrieval and peak retrieval, respectively.

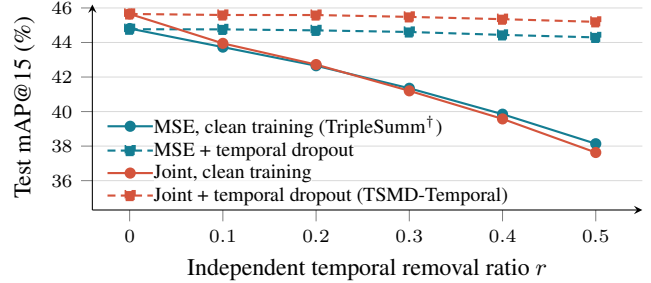
\begin{figure}[t]
\centering
\begin{tikzpicture}
\begin{axis}[
    width=1.03\columnwidth, height=0.50\columnwidth,
    xlabel={Independent temporal removal ratio $r$},
    ylabel={Test mAP@15 (\%)},
    xmin=-0.04, xmax=0.55, ymin=34.3, ymax=46.2,
    xtick={0,0.1,0.2,0.3,0.4,0.5},
    ytick={36,38,40,42,44,46},
    ymajorgrids, grid style={gray!30, line width=0.3pt},
    axis lines=left,
    tick align=outside, tick pos=left,
    label style={font=\small}, tick label style={font=\scriptsize},
    legend columns=1, legend cell align=left,
    legend style={at={(0.015,0.0)}, anchor=south west, draw=none,
              font=\fontsize{7.5}{7.5}\selectfont,
              row sep=2pt,
              inner sep=2pt, fill=none,
              nodes={fill=white, fill opacity=0.7,
                     text opacity=1, inner sep=0pt}}, 
]
\addplot[MColor, thick, mark=*, mark size=1.7pt]
  coordinates {(0,44.809)(0.1,43.732)(0.2,42.654)(0.3,41.349)(0.4,39.848)(0.5,38.137)};
\addlegendentry{MSE, clean training (TripleSumm$^\dagger$)}
\addplot[MColor, thick, densely dashed, mark=square*, mark size=1.7pt]
  coordinates {(0,44.768)(0.1,44.759)(0.2,44.702)(0.3,44.609)(0.4,44.440)(0.5,44.295)};
\addlegendentry{MSE + temporal dropout}
\addplot[JointColor, thick, mark=*, mark size=1.7pt]
  coordinates {(0,45.655)(0.1,43.949)(0.2,42.722)(0.3,41.201)(0.4,39.575)(0.5,37.630)};
\addlegendentry{Joint, clean training}
\addplot[JointColor, thick, densely dashed, mark=square*, mark size=1.7pt]
  coordinates {(0,45.653)(0.1,45.593)(0.2,45.590)(0.3,45.483)(0.4,45.350)(0.5,45.196)};
\addlegendentry{Joint + temporal dropout (TSMD-Temporal)}
\end{axis}
\end{tikzpicture}
\vspace{-10pt}
\caption{MoSu test mAP@15 under increasing independent temporal removal ratio. Blue and red curves use the MSE-only and joint objectives. Solid and dashed curves denote clean and temporal-dropout training, respectively. Their separation shows the complementary contributions of the joint objective and temporal-dropout training.}
\label{fig:frame_drop}
\end{figure}

Robustness is benchmarked under two regimes. Temporal removal is reported at $r=0.5$ in Table~\ref{tab:main}(b) and swept over $r \in \{0, 0.1, \dots, 0.5\}$ in Fig.~\ref{fig:frame_drop}. Stream removal is reported in Table~\ref{tab:main}(c) as the mean over $-V$/$-A$/$-T$, with a per-stream breakdown in Table~\ref{tab:per_modality}. The two regimes provide reciprocal out-of-distribution tests: TSMD-Temporal is evaluated under stream removal, and TSMD-Stream under temporal removal. For temporal removal, each checkpoint is evaluated with three fixed mask seeds. Values are averaged across the three training seeds, and $\pm$ denotes their SD.

\begin{table}[t]
\centering
\caption{Three-seed mean results for objective components under clean inputs, where M, P, and R denote MSE, Pearson, and RankNet losses, respectively.}
\label{tab:objective_ablation}
\setlength{\tabcolsep}{2.5pt}
\begin{tabular}{l ccc ccc}
\toprule
\multirow{2}{*}{\textbf{Objective}} & \multicolumn{3}{c}{\textbf{MoSu}} & \multicolumn{3}{c}{\textbf{Mr.~HiSum}} \\
\cmidrule(lr){2-4} \cmidrule(lr){5-7}
& $\tau$ & $\rho$ & mAP@15 & $\tau$ & $\rho$ & mAP@15 \\
\midrule
M       & 0.353 & 0.473 & 44.81 & 0.255 & 0.347 & 41.12 \\
M+P     & \textbf{0.361} & \textbf{0.481} & 45.59 & \textbf{0.263} & \textbf{0.356} & 41.64 \\
M+R     & 0.353 & 0.475 & \textbf{45.82} & 0.249 & 0.341 & \underline{41.81} \\
M+P+R   & \underline{0.360} & \underline{0.481} & \underline{45.65} & \underline{0.260} & \underline{0.353} & \textbf{42.31} \\
\bottomrule
\end{tabular}
\end{table}

\begin{table}[t]
\centering
\vspace{-10pt}
\caption{Three-seed mean test mAP@15 under each missing stream using the joint objective.}
\label{tab:per_modality}
\setlength{\tabcolsep}{4pt}
\begin{tabular}{@{}ll ccc@{}}
\toprule
Dataset & Training policy & $-V$ & $-A$ & $-T$ \\
\midrule
\multirow{2}{*}{MoSu}
& Clean training & 36.37 & 38.99 & 42.60 \\
& $+$ TSMD-Stream & \textbf{42.07} & \textbf{43.37} & \textbf{44.46} \\
\midrule
\multirow{2}{*}{Mr.~HiSum}
& Clean training & 38.90 & 36.77 & \textbf{41.77} \\
& $+$ TSMD-Stream & \textbf{41.03} & \textbf{38.62} & 41.55 \\
\bottomrule
\end{tabular}
\vspace{-3pt}
\end{table}

\vspace{-2pt}
\subsection{Results and Discussion}
\label{ssec:results}

\subsubsection{Clean-Input Performance and Objective Selection}
Table~\ref{tab:objective_ablation} evaluates the joint objective under clean inputs. The M row (TripleSumm$^\dagger$ in Table \ref{tab:main}) reports our three-seed retraining of TripleSumm with its original MSE-only objective ($L_M$). It closely reproduces the published mAP@15 results ($44.81$ vs.\ $44.42$ on MoSu; $41.12$ vs.\ $40.88$ on Mr.~HiSum) and gives the lowest mAP@15 among the evaluated objectives on both datasets. Incorporating per-video Pearson loss ($L_P$) enforces sequence-level profile alignment, yielding improvements in $\tau$ and $\rho$ across both datasets ($+0.008 / +0.008$ on MoSu; $+0.008 / +0.009$ on Mr.~HiSum). Conversely, adding peak-oriented RankNet loss ($L_R$) concentrates gradient updates on salient moments, boosting peak retrieval ($45.82$ and $41.81$ mAP@15).

Unifying all three terms, the joint objective achieves the second-highest correlation on MoSu while securing the top peak retrieval score on Mr.~HiSum ($42.31$ mAP@15). Comparing the TSMD variants in Table~\ref{tab:main}(a) with the M+P+R row of Table~\ref{tab:objective_ablation}, which shares the same objective but uses clean training, all variants match or exceed it on MoSu ($45.65$ mAP@15) and remain within $0.25$ points of it on Mr.~HiSum ($42.31$). Thus, training with modality dropout does not substantially sacrifice clean-input performance.

\subsubsection{Robustness Under Temporal Removal}
Under 50\% independent temporal removal (Table~\ref{tab:main}(b)), TripleSumm$^\dagger$ decreases by 6.67 mAP@15 on MoSu ($44.81 \rightarrow 38.14$) and 2.93 on Mr.~HiSum ($41.12 \rightarrow 38.19$), accompanied by a reduction in $\tau$ of 0.102 and 0.076. TSMD-Temporal reaches 45.20 and 41.60 mAP@15 (+7.06 and +3.41 over TripleSumm$^\dagger$) while maintaining $\tau$ within 0.01 of its clean values. As illustrated in Fig.~\ref{fig:frame_drop}, at the evaluated removal ratios, clean-trained models consistently degrade as $r$ increases, whereas temporal-dropout training limits the total mAP@15 reduction from $r=0$ to $r=0.5$ to less than 0.5 points under both objectives. The joint objective also maintains a 0.83--0.91 point advantage over MSE-only training throughout the temporal-dropout sweep.

\subsubsection{Stream Removal and Cross-Regime Complementarity}

Under complete-stream removal (Table~\ref{tab:main}(c)), TSMD-Stream ranks first on all eight metrics, improving mAP@15 over TripleSumm$^\dagger$ by $4.41$ on MoSu and $2.59$ on Mr.~HiSum. Each specialized policy also transfers partially to the other regime: TSMD-Temporal still gains $3.45$ and $1.51$ mAP@15 under stream removal, and TSMD-Stream gains $4.32$ and $0.88$ under temporal removal. However, each trails the matched specialist by up to $1.08$ and $2.74$ points, respectively, indicating that robustness is strongest when the scale of training corruption matches the test-time failure.

Because its training covers both failure scales, TSMD-Mix closes most of these gaps. Under temporal removal, it matches TSMD-Temporal, with mAP@15 differences of at most $0.13$. Under stream removal, it ranks second on all metrics and remains within $0.24$ mAP@15 of TSMD-Stream. Thus, halving the exposure to each failure type costs little robustness, and TSMD-Mix provides a robust compromise when the failure type is unknown at deployment.

Per-modality breakdowns (Table~\ref{tab:per_modality}) show dataset-specific failure modes: under the clean-trained joint objective, MoSu is most affected by visual absence ($36.37$ mAP@15), whereas Mr.~HiSum is most affected by audio absence ($36.77$). TSMD-Stream raises these respective worst cases to $42.07$ and $38.62$ and improves the average across missing streams on both datasets, with only a marginal decrease on Mr.~HiSum $-$T ($41.77 \rightarrow 41.55$).

%% file: section/4.Conclusion.tex
\section{Conclusion}
\vspace{-4pt}
\label{sec:conclusion}
We studied multimodal missingness in video highlight detection along temporal and stream-level dimensions. Across the evaluated conditions on MoSu and Mr.~HiSum, robustness of the proposed framework is strongest when the scale of training corruption matches the test-time failure. TSMD-Temporal is most effective for localized frame missingness, whereas TSMD-Stream is strongest when an entire stream is unavailable. TSMD-Mix remains the best or the second-best across the evaluated temporal and stream-level conditions without changing the model architecture or inference procedure. Future work will extend beyond zero-masking to realistic degradations such as feature noise, transcript errors, and cross-stream misalignment.